# A Possibilistic Model for Qualitative Sequential Decision Problems under Uncertainty in Partially Observable Environments

Régis Sabbadin
IRIT - Université Paul Sabatier - 31062 Toulouse Cedex (France)
e-mail: sabbadin@irit.fr

## Abstract

In this article we propose a qualitative (ordinal) counterpart for the Partially Observable Markov Decision Processes model (POMDP) in which the uncertainty, as well as the preferences of the agent, are modeled by possibility distributions. This qualitative counterpart of the POMDP model relies on a possibilistic theory of decision under uncertainty, recently developed.

One advantage of such a qualitative framework is its ability to escape from the classical obstacle of stochastic POMDPs, in which even with a finite state space, the obtained *belief state* space of the POMDP is infinite. Instead, in the possibilistic framework even if exponentially larger than the state space, the belief state space remains finite.

## 1 INTRODUCTION

The Partially Observable Markov Decision Processes model (POMDP) is a general model for sequential decision problems in which effects of actions as well as results of observations are noisy, and the noise is represented by probability distributions.

The POMDP model takes its justification from the most widely used theory for decision under uncertainty : the *expected utility theory* [Savage1954] [von Neumann & Morgenstern1944]. It is theoretically very attractive but its practical use remains difficult, especially because it involves infinite state spaces.

In this article we propose an ordinal counterpart of the POMDP model in which uncertainty as well as preferences are modeled by qualitative possibility distributions that take their values in a finite ordinal scale. The underlying decision criterion is based on the use of the Sugeno integral which is an ordinal, *maxmin*, counterpart of the Choquet integral generalizing the expected utility criterion. It has received axiomatic justifications in the style of von Neumann and Morgenstern [Dubois & Prade1995] and recently in the style of Savage [Dubois, Prade, & Sabbadin1998].

In the next section we will describe the usual framework of fully and partially observable markov decision processes as well as some classical resolution methods. We will recall, in particular, how POMDPs can be restated as fully observable MDPs on the (infinite, continuous) belief state space, thus leading to define methods that approximate the value of optimal policies, either by updating a value function, or by discretizing the continuous belief state space.

Then, in section 3 we will give some background on the possibilistic utility functions advocated by [Dubois & Prade1995], and we will see how [Fargier, Lang, & Sabbadin1998] proposed to extend them to multistage decision making (in a fully observable domain). In particular, we will recall some decomposability properties that allow to use dynamic programming-like algorithms to solve such possibilistic multistage decision problems (that we call Π-MDPs). In [Fargier, Lang, & Sabbadin1998], especially, a backwards induction algorithm is proposed, for solving finite-horizon Π-MDPs. In the present article, we first give a slightly different version of this algorithm, which makes it similar to the *value iteration* algorithm used for solving classical MDPs, and allows to handle the infinite horizon as well as the finite horizon case.

Section 4 is the central part of the paper : here we will recall how conditioning can be defined (not uniquely, by the way) in possibility theory. We will then define a possibilistic counterpart of POMDPs which, we will show, can be handled by finite domain Π-MDP algorithms, even if considering partial observability in Π-POMDPs makes the computation of optimal policies exponentially (in space and time) more difficult than in the completely observable case.

## 2 MARKOV DECISION PROCESSES

### 2.1 FULLY OBSERVABLE MARKOV DECISION PROCESSES

The standard MDP model [Puterman1994] is defined by :

- A set $T \subseteq \mathbb{N}$ of stages in which decisions must be taken. When $T$ is finite ($T = \{0, \ldots N\}$), $N$ is the horizon of the problem.
- For each stage $t$, a finite state space, $S_t$.
- Sets $A_{s,t}$ (finite) of available actions in state $s$ at stage $t$ (these sets are denoted $A_s$ when they do not depend on $t$).
- The rewards $r(s, a)$ that are obtained after $a$ has been applied in state $s$. These rewards may be negative, thus considered as costs or penalties.
- To each action $a \in A_{s,t}$ applied in state $s \in S_t$ is assigned a probability distribution $p(\cdot|s, a)$ describing the uncertainty about the possible successor states in $S_{t+1}$.

A *decision rule* $d_t$ is an application from $S_t$ to $\cup_{s \in S_t} A_{s,t}$ assigning an action to each possible state of the world in stage $t$. The set of these decision rules is denoted by $D_t$. A *policy* $\delta$ is, in the finite horizon case, a n-tuple of decision rules $\delta = (d_1, \ldots, d_N)$ where $N$ is the horizon of the problem. $\Delta = D_1 \times \ldots \times D_N$ is the set of applicable policies. In the infinite horizon case, or in the *stationary* finite horizon case, the parameter $t$ has no influence on the decision problem. Thus, a policy $\delta$ is nothing but the repetition of an identical decision rule $d$.

A policy $\delta$, applied in an initial state $s_0$, defines a *Markov chain* that describes the sequence of states occupied by the system (the trajectory $\tau = \{s_0, \ldots, s_N\}$). The *value of a policy* in a given state is the expected sum of the rewards gained along the trajectory. In the finite horizon case, it is :

$$v(\delta, s_0) = E(\sum_{t=0}^{N} r(s_t, d_t(s_t))) \quad (1)$$

When the horizon is infinite, the above expected sum may be unbounded. Therefore, future rewards are usually discounted, which is in accordance with the fact that immediate rewards shall be more important than future ones. In this case, the discounted value of a policy is defined by :

$$v(\delta, s_0) = E(\sum_{t=0}^{\infty} \gamma^t \cdot r(s, d_t(s))) \quad (2)$$

where $0 < \gamma < 1$ is the discounting factor (the sum converges, since $\gamma < 1$).

Solving a MDP amounts to finding a policy $\delta^*$ maximizing $v(\cdot, s_0)$. The *dynamic programming* methods [Puterman1987] are based on the decomposition of the sequential decision problem into one-stage decision problems, by making use of the Bellman's equations [Bellman1957].

In the finite horizon case, an optimal policy for an MDP is obtained as the solution of the following system of equations :

$\forall t \in 0, \ldots, N-1, \forall s \in S_t,$

$$v^t(s) = \max_{a \in A_{s,t}} \{r(s, a) + \sum_{s' \in S_{t+1}} p_a(s'|s) \cdot v^{t+1}(s')\} \quad (3)$$

$$v^N(s) = \max_{a \in A_s} r(s, a)$$

Optimal policies can be computed by the *backwards induction* algorithm [Puterman1987], which solves the above equations in decreasing order of $t$.

**Algorithm 1:** Backwards induction

**begin**
$\quad t \leftarrow N + 1$ ;
$\quad \forall s, v^{N+1}(s) \leftarrow 0$ ;
$\quad$ **while** $t \geq 1$ **do**
$\quad\quad t \leftarrow t - 1$ ;
$\quad\quad$ **for** $s \in S_t$ **do**
$\quad\quad\quad v^t(s) \leftarrow \max_{a \in A_{s,t}} \{r(s, a) + \sum_{s' \in S_{t+1}} p(s'|s, a) \cdot v^{t+1}(s')\}$ ;
$\quad\quad\quad D_t^*(s) = argmax_{a \in A_{s,t}} \{r(s, a) + \sum_{s' \in S_{t+1}} p(s'|s, a) \cdot v^{t+1}(s')\}$ ;
$\quad$ **return** $v^0(s_0), \{\delta^*\}$
**end**

$D_t^*(s)$ is the set of optimal actions in state $s$ and stage $t$, $\{\delta^*\}$ is the set of policies such that $\forall t \in 0, \ldots, N$, $\forall s \in S_t, d_t^*(s) \in D_t^*(s)$.

In the discounted infinite horizon case, optimal policies can be obtained as fixed points of equation (3). Methods such as the *value iteration* algorithm [Bellman1957], [Bertsekas1987] can be used to compute these optimal policies which furthermore, are stationary.

In the value iteration algorithm, the function $Q^*(s, a)$ represents the value of performing action $a$ in state $s$. It is used instead of $v(s)$, which is the value of performing the optimal action in state $s$. $Q^*(s, a)$ is defined by

$$Q^*(s, a) = r(s, a) + \gamma \cdot \sum_{s' \in S} p(s'|s, a) \cdot v(s') \quad (4)$$

**Algorithm 2:** Value iteration

**begin**

   Arbitrary initialization of $v$ on $S$ ;

   **repeat**

      **for** $s \in S$ **do**

         **for** $a \in A$ **do** $Q(s,a) \leftarrow r(s,a) + \gamma \cdot \sum_{s' \in S} p(s'|s,a) \cdot v(s')$ ;

         $v(s) \leftarrow \max_a Q(s,a)$ ;

   **until** $Q$ converges to $Q^*$;

   return $Q^*$

**end**

and $\forall s \in S, v(s) = \max_{a \in A_s} Q^*(s,a)$.

Results about the convergence of algorithm 2 can be found in [Bertsekas1987]. It is easy to get an optimal policy $\delta^*$ from $Q^*$, since $\delta^*(s) = argmax_a Q^*(s,a)$.

Many other algorithms have been designed to solve infinite horizon MDPs, a review of which can be found in [Puterman1994].

### 2.2 PARTIALLY OBSERVABLE MARKOV DECISION PROCESSES

POMDPs [Monahan1982] [Lovejoy1991] are a generalization of MDPs in which it is not assumed that the agent knows precisely the state $s$ of the system, in each decision stage. Its imprecise belief is modeled by a *belief state* $b$, which is a probability distribution on the state space $S$, regularly updated by *observations*.

The observation model consists of an *observation set* $O$ of possible observations and a set of probability distributions over $O$, $p(\cdot|s,a)$, where, for all $o \in O$, $p(o|s,a)$ is the probability of observing $o$ after $a$ was applied and the resulting state of the system is $s$.

The usual MDP techniques cannot be applied directly to a POMDP since they assume that the current state of the system, $s$, is always known. A way to solve a POMDP is to assume that it is a MDP over the *belief state space*. After performing action $a$ in belief $b$, the agent does not know the precise state of the system, but it can compute a resulting belief state $b_a$, that it can update to $b_a^o$ when it observes $o$.

[Cassandra, Kaelbling, & Littman1994] gave the following equations, linking $b$, $b_a$ and $b_a^o$ :

$$b_a(s) = \sum_{s' \in S} p(s|s',a) \cdot b(s') \quad (5)$$

$$b_a(o) = \sum_{s \in S} p(o|s,a) \cdot b_a(s) \quad (6)$$

$$b_a^o(s) = \frac{p(o|s,a) \cdot b_a(s)}{b_a(o)} \text{ when } b_a(o) \neq 0 \quad (7)$$

$b_a(o)$ is the probability of observing $o$ when performing $a$ in $b$. In other terms, $b_a(o)$ is the probability of being in belief state $b_a^o$ in these conditions : $b_a(o) = p(b_a^o|b,a)$. It is indeed a transition probability in a new MDP relating belief states.

The Bellman equation of this new MDP is the following :

$$v^t(b) = \max_{a \in A_b}\{r(b,a) + \sum_{o \in O} b_a(o) \cdot v^{t+1}(b_a^o)\} \quad (8)$$

where $r(b,a) = \sum_{s \in S} r(s,a) \cdot b(s)$ is the average reward that can be collected in $b$ and $A_b$ is the set of the actions that are available in at least one possible state.

The obtained MDP has a continuous state space which makes its resolution difficult (PSPACE complete [Bertsekas & Tsitsiklis1989]).

[Cassandra, Kaelbling, & Littman1994] proposed to construct an approximation $\hat{v}^t$ of $v^*$ (the fixed point of (8)) and to improve it over time steps. Others, like [Parr & Russel1995], [Brafman1997] or [Hauskrecht1997] propose to discretize $B_t$ and compute optimal policies over the obtained finite subset of the belief space.

## 3 POSSIBILISTIC FULLY OBSERVABLE MULTISTAGE DECISION

### 3.1 POSSIBILISTIC DECISION CRITERIA

[Dubois & Prade1995] proposed an ordinal counterpart, based on possibility theory, of the expected utility theory for one-stage decision making. In this framework, $S$ and $X$ are respectively the (finite) sets of possible states of the world and consequences of actions. $L$ is a finite totally ordered (qualitative) scale, which lowest and greatest elements are denoted respectively $0_L$ and $1_L$.

The uncertainty of the agent about the effect of an action $a$ taken in state $s$ is represented by a possibility distribution $\pi(\cdot|s,a) : X \to L$. $\pi(x|s,a)$ measures to what extent $x$ is a plausible consequence of $a$ in $s$. $\pi(x|s,a) = 1_L$ means that $x$ is completely plausible, whereas $\pi(x|s,a) = 0_L$ means that it is completely impossible.

In the same way, consequences are ordered in terms of level of satisfaction by a qualitative utility function $\mu : X \to L$. $\mu(x) = 1_L$ means that $x$ is completely satisfactory, whereas if $\mu(x) = 0_L$, it is totally unsatisfactory. Notice that $\pi$ is normalized (there shall be at least one completely possible state of the world), but $\mu$ may not be (it can be that no consequence is totally satisfactory).

[Dubois & Prade1995] proposed the two following qualitative decision criteria :

$$u^*(a, s_0) = \max_{x \in X} \min\{\pi(x|s_0, a), \mu(x)\} \quad (9)$$

$$u_*(a, s_0) = \min_{x \in X} \max\{n(\pi(x|s_0, a)), \mu(x)\} \quad (10)$$

where $n$ is the order reversing map of $L$.

Note that $u^*(a, s_0)$ corresponds to the degree of intersection of the fuzzy set of plausible consequences of $a$ in $s_0$ (which membership function is $\pi(\cdot|s_0, a)$) and the fuzzy set of preferred consequences (which membership function is $\mu$). $u_*(a, s_0)$ is instead a degree of inclusion of the first fuzzy set into the second.

$u^*$ can be seen as an extension of the *maximax* criterion which assigns to an action the utility of its best possible consequence. On the other hand, $u_*$ is an extension of the *maximin* criterion which corresponds to the utility of the worst possible consequence. $u_*$ measures to what extent every plausible consequence is satisfactory.

$u^*$ corresponds to a very adventurous (optimistic) attitude in front of uncertainty, whereas $u_*$ is conservative (cautious). We will focus on the second criterion in the rest of the paper, as we will have some guarantee that the conservative-optimal policies eventually lead to some non totally unsatisfactory state (we have no similar guarantee with adventurous optimal policies).

### 3.2 Π-MDP : THE FINITE-HORIZON CASE

In [Fargier, Lang, & Sabbadin1998], the possibilistic qualitative decision theory has been extended to finite-horizon, multistage decision making.

In this framework, the qualitative (conservative) utility of a policy $\delta$ in state $s_0$ is defined by the qualitative expectation (*minmax*) of the *minimum* of the degrees of satisfaction of the states of the possible trajectories :

$$u_*(\delta, s_0) = \min_\tau \max\{n(\pi(\tau|s_0, \delta)), \mu(\tau)\} \quad (11)$$

where, if $\tau = \{s_0, \ldots, s_N\}$, $\mu(\tau) = \min_{i \in 0 \ldots N} \mu(s_i)$ and $\pi(\tau|s_0, \delta) = \min_{i \in 0 \ldots N-1} \pi(s_{i+1}|s_i, d_i(s_i))$.

The possibilistic counterpart of the Bellman equation is the following:

$$u_*^t(s) = \max_{a \in A_s} \min\{\mu(s), \min_{s' \in S_{t+1}} \max\{n(\pi(s'|s,a)), u_*^{t+1}(s')\}\}$$

$$u_*^N(s) = \mu(s) \quad (12)$$

In [Fargier, Lang, & Sabbadin1998] it is shown that any policy computed backwards by successive applications of (12) is optimal according to $u_*$. The following algorithm performs such a computation (in the case where intermediate satisfaction degrees are not considered) :

**Algorithm 3:** Possibilistic backwards induction

**begin**
$\quad t \leftarrow N$ ;
$\quad \forall s, u_*^N(s) \leftarrow \mu(s)$ ;
$\quad$ **while** $t \geq 1$ **do**
$\quad\quad t \leftarrow t - 1$ ;
$\quad\quad$ **for** $s \in S_t$ **do**
$\quad\quad\quad u_*^t(s) \leftarrow \max_{a \in A_s} \min_{s' \in S_{t+1}} \{n(\pi(s'|s, d_t(s))), u_*^{t+1}(s')\}$ ;
$\quad\quad\quad D_t^*(s) \leftarrow argmax_{a \in A_s} \min_{s' \in S_{t+1}} \{n(\pi(s'|s, d_t(s))), u_*^{t+1}(s')\}$ ;

$\quad$ **return** $u_*^0(s_0), \{\delta^* \in D_0^*\}$
**end**

Note that because of the idempotency of the *minimum* operator, there are optimal policies that may not be found by such an algorithm, unlike in the stochastic case. Anyway, every policy returned by the algorithm is optimal, and has the property that any subpolicy applied from stage $t$ to $N$ (the horizon) is optimal (see [Fargier, Lang, & Sabbadin1998]).

### 3.3 Π-MDP : A VALUE ITERATION ALGORITHM

Let us now change a bit the data of the problem, in order to recover one that admit stationary optimal policies in the infinite horizon case. First of all, suppose that the state spaces, the available actions and the transition functions do not depend on the stage of the problem. Suppose also that a utility function $\mu$ on $S$ is given, that expresses the preferences of the agent on the states that the system shall reach and stay in. We finally assume the existence of an action *stay*, that keeps the system in the same state (or equivalently, an action $do - nothing$, if we assume that the system does not evolve by itself, without any action applied).

Then, under these assumptions, we are able to define a possibilistic counterpart of the *value iteration* algorithm, that computes optimal policies from iterated modifications of a possibilistic value function.

First, we have to define $\tilde{Q}^*$, the possibilistic counterpart of $Q$-functions. As in the stochastic case, $\tilde{Q}^*(s, a)$ evaluates the "utility" of performing $a$ in $s$. We have a similar property as in the stochastic case, that is that the optimal possibilistic strategy can be obtained from the solution of the following equations :

**Proposition 1** *The optimal strategy can be obtained*

*from $\tilde{Q}^*$ the solution of the following set of equations :*

$$\tilde{Q}^*(s,a) = \min_{s' \in S} \max\{n(\pi(s'|s,a)), u_*(s')\},$$

*where $u_*(s) = \max_a \tilde{Q}^*(s,a)$.*

Then, we define a possibilistic version of the value iteration algorithm that computes $\tilde{Q}^*$ : the *possibilistic value iteration algorithm* (algorithm 4).

**Algorithm 4:** Possibilistic value iteration

**begin**
    $u_*(s) = \mu(s), \forall s \in S$ ;
    **repeat**
        **for** $s \in S$ **do**
            **for** $a \in A$ **do**
                $\tilde{Q}(s,a) \leftarrow$
                $\min_{s' \in S} \max\{n(\pi(s'|s,a)), u_*(s')\}$ ;
            $u_*(s) \leftarrow \max_a \tilde{Q}(s,a)$ ;
    **until** $\tilde{Q}$ converges to $\tilde{Q}^*$;
    **return** $\tilde{Q}^*$
**end**

This algorithm converges to the actual value of $\tilde{Q}^*$ in a finite number of step. This is easy to show, once it is noticed that the sequence of functions ($\tilde{Q}^*$) computed by the algorithm takes its values in a finite set and is non-decreasing. The number of iteration is, by the way, bounded by the size of the set of possible value functions : $|A| \times |S| \times |L|$. As one iteration of the algorithm requires $|S| \times |A|$ evaluations of $\tilde{Q}(s,a)$, the overall complexity of finding an optimal policy is in $O(|S|^2 \times |A|^2 \times |L|)$.

Notice that unlike in the stochastic value iteration algorithm, the initialization of $u_*$ is not arbitrary. Notice also, that after $k$ iterations of the algorithm, the $\tilde{Q}$-function allows to determine the "best" policies that allow to reach goal states in no more than $k$ actions. Moreover, the policies that are computed are not only the best according to the pessimistic utility, but among the best, they are those that guarantee a shortest path to a goal state.

**Example**

A robot is located somewhere in a room. The point is to define a policy that is able to bring it into the down-right square of the room shown in Figure 1. The objective will be partially satisfied if the robot ends in one of the neighbor squares. The state-space and the utility function $\mu$ on the objective states (taking its values in a finite subset of the interval $[0,1]$) are depicted in Figure 1. $\mu(s_{33}) = 1$, $\mu(s_{23}) = \mu(s_{32}) = 0.5$ and $\mu(s) = 0$ for the other states.

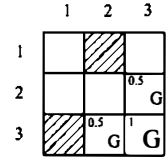

Figure 1: State space and utility function.

The available actions are to move (T)op, (D)own, (L)eft, (R)ight or to (S)tay in place. If the robot chooses to stay, it will *certainly* remain in the same square. If it goes T, D, L or R it will (entirely) possibly reach the desired square ($\pi = 1$) if it is free but it will be possible that it reaches a neighbor square, as depicted in Figure 2 for the action R. The other transition possibility functions are of course symmetric to these.

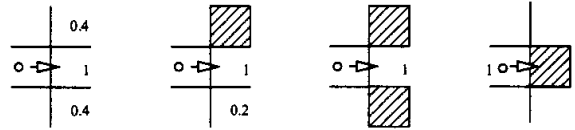

Figure 2: Transition possibilities for moving right.

Let us now compute the optimal actions after one iteration of algorithm 4. For every action $a$ and state $s$, we have $\tilde{Q}^1(s,a) = \min_{s' \in S} \max(1 - \pi(s'|s,a), \mu(s'))$ ($\pi$ does not depend on the stage) and $u_*^1(s) = \max_{a \in \{T,D,L,R,S\}} \tilde{Q}^1(s,a)$.

Figure 3 Resumes the utility of each state after one iteration, as well as an action that is optimal if the problem is assumed to be solved in one iteration only, for each state with a non-null pessimistic utility. The optimal action is unique, except for state $s_{33}$ for which $D$ and $R$ would be optimal actions as well.

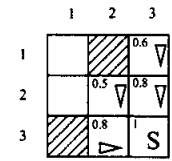

Figure 3: Optimal policy, computed at iteration 1.

Now we can iterate the process and get an optimal policy. The iterated process is described in Figure 4. Note that after 4 iterations, the utility of each state and the associated optimal action do not change anymore.

The number of iterations required to find a stationary policy is about the number of steps of the longest path from any state to goal state $S$, actions being assumed to be deterministic.

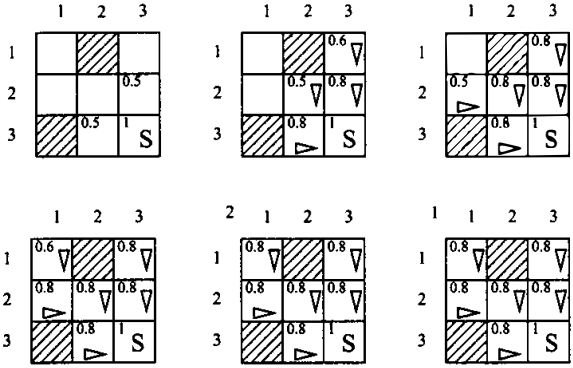

Figure 4: Pessimistic optimal policy computation.

## 4 POSSIBILISTIC PARTIALLY OBSERVABLE MULTISTAGE DECISION

### 4.1 BACKGROUND ON POSSIBILISTIC CONDITIONING

Conditioning has been defined in possibility theory (see [Dubois & Prade1994] for a complete presentation). As applied to events, possibilistic conditioning shall obey a form which is similar to the one of Bayesian conditioning :

$$\forall A, B, \Pi(A \cap B) = \Pi(B|A) * \Pi(A) \qquad (\alpha)$$

where $*$ can be *min* (see [Hisdal1978]) or *product*.

As the framework we choose to consider is purely ordinal, the only choice left is the operator *min*. Unlike in Bayesian conditioning, equation $(\alpha)$ does not have a unique solution $\Pi(B|A)$. It is common then, after [Dubois & Prade1990], to choose the solution to $(\alpha)$ which has the *minimum specificity*, thus leading to define $\Pi(B|A)$ as :

$\Pi(B|A) = 1_L$ if $\Pi(A \cap B) = \Pi(A) > 0_L$ and $\Pi(B|A) = \Pi(A \cap B)$ otherwise.

Once the conditioning of possibility measures has been defined in this way, the conditioning of a possibility distribution by an observation $o \in O$ ($\pi(\cdot|o)$) can be defined as :

**Definition 1** *Conditional possibility distribution :* $\pi(s|o) = 1_L$ *if* $\pi(s,o) = \Pi(o)$ *and* $\pi(s|o) = \pi(s,o)$ *otherwise. Where* $\Pi(o) = max_s \pi(s,o)$ *and* $\pi(\cdot, \cdot)$ *is a joint possibility distribution over* $S \times O$.

### 4.2 POSSIBILISTIC POMDPs

We have already seen how classical POMDPs can be translated into MDPs over a *belief state space*. In a similar way, we are going to see how possibilistic POMDPs (Π-POMDPs) can be translated into Π-MDPs over a *possibilistic belief state space*.

First of all, let us define a possibilistic belief state, $\beta$ as a possibility distribution over the state space, expressing a plausibility ordering over the states. Unlike the set of probabilistic belief states, the set of possibilistic beliefs is finite, as soon as we assume that the possibility degrees take their values in a finite set $L$ (the cardinal of the possibilistic belief state space $B$ is bounded by $|S| \times |L|$).

Suppose now that as in the probabilistic case, the transition possibilities $\pi(s|s', a)$ are given, as well as the observation possibilities $\pi(o|s, a)$. Similarly to [Cassandra, Kaelbling, & Littman1994] we can define $\beta_a(s)$ as the possibility of reaching state $s$, from an initial belief $\beta$ and after the execution of action $a$ :

$$\beta_a(s) = \max_{s' \in S} \min\{\pi(s|s', a), \beta(s')\} \qquad (13)$$

We can also compute the possibility of observing $o \in O$ after having applied $a$ in the (possibilistic) belief state $\beta$ :

$$\beta_a(o) = \max_{s \in S} \min\{\pi(o|s, a), \beta_a(s)\} \qquad (14)$$

Now, $\beta_a^o$ is the revised possibilistic belief state, after $a$ was applied in $\beta$ and $o$ was observed :

- $\beta_a^o(s) = 0_L$ if $\pi(o|s, a) = 0_L$,
- $\beta_a^o(s) = 1_L$ if $\pi(o|s, a) = \beta_a(o) > 0_L$, (15)
- $\beta_a^o(s) = \beta_a(s)$ in the other cases.

Then, all the elements of the new Π-MDP over possibilistic belief states are defined in equations (13), (14) and (15). The new Π-MDP may be intuitively expressed as : from a state $\beta$, applying action $a$ may lead to one of the $|O|$ states $\beta_a^o$, the possibility of reaching $\beta_a^o$ being $\beta_a(o) = \pi(o|\beta, a)$.

From now on, the "possibilistic" Bellman equation (pessimistic case) can be extended to the partially observable case :

$$u_*^t(\beta) = \max_{a \in A_*} \min\{\mu(\beta), \min_{o \in O} \max\{n(\beta_a(o)), u_*^{t+1}(\beta_a^o)\}\}$$
(16)

where $\mu(\beta) = \min_{s \in S} \max\{n(\beta(s)), \mu(s)\}$ and $u_*^0(\beta)$ is initialized to $\mu(\beta)$.

In the same way, the $\tilde{Q}$ function becomes :

$$\tilde{Q}(\beta, a) = \min_{o \in O} \max\{n(\beta_a^o), u_*(\beta_a^o)\} \qquad (17)$$

Algorithm 4 can now be extended to the partially observable case (in case no intermediate satisfaction degree is involved).

**Algorithm 5:** Possibilistic value iteration in the partially observable case

**begin**

$u_*(\beta) = \mu(\beta), \forall \beta \in B$ ;
**repeat**
  **for** $\beta \in B$ **do**
    **for** $a \in A$ **do**
      $\tilde{Q}(\beta, a) \leftarrow \min_{o \in O} \max\{n(\beta_a(o)), u_*(\beta_a^o)\}$ ;
      $u_*(\beta) \leftarrow \max_a \tilde{Q}(\beta, a)$ ;
**until** $\tilde{Q}$ *converges to* $\tilde{Q}^*$;
**return** $\tilde{Q}^*$

**end**

Here, $B$ is the set of belief states, which size is at most $|L|^{|S|}$. For a given $\beta$ and a given $a$, evaluating $\tilde{Q}(\beta, a)$ needs $|O|$ computations of $\beta_a^o$, which needs about $|S|$ elementary computations (as $\beta_a^o$ is immediately deduced from $\beta_a(o)$). The algorithm will perform at most $|B| \times |L| \times |A|$ iterations before convergence. Finally, the overall complexity of the algorithm is in $O(|L|^{2 \cdot |S|+1} \times |A|^2 \times |O|)$. This worst-case complexity is of course exponential in the size of the state space, in the general case, but it can be reduced when the subset of belief states that can be reached from an initial belief state is small.

**Example**

Le us take again the preceding example, but assume now that observability is no more complete : all what the robot knows when it is in a given state (square) is the configuration of the walls around the square. Initially, the robot observes nothing, that is it can be anywhere in the state space depicted in figure 1.

Let us assume also that the observations are not noisy, then it can be shown that the belief states that can be reached from the initial belief state, applying any policy are limited to the following set of subsets of $S$ : $B = \{S, \{s_{21}, s_{32}\}, \{s_{22}\}, \{s_{23}\}, \{s_{32}\}, \{s_{33}\}, \{s_{11}, s_{13}\}, \{s_{21}\}\}$. These 8 belief states will be denoted by $\beta_0$ to $\beta_7$. We can compute $\mu(\beta_i)$ from the utility function over $S$ : $\mu(\beta_0) = \mu(\beta_1) = \mu(\beta_2) = \mu(\beta_6) = \mu(\beta_7) = 0$, $\mu(\beta_3) = \mu(\beta_4) = 0.5$ and $\mu(\beta_5) = 1$.

The possibilities of transition from any belief state to any other, given an action $a$ can be computed from the $\beta_a(o)$'s and the $\beta_a^o$'s, for $o \in O$. For example, if the initial belief state is $\beta_0 = S$, and the action that is performed is (D)own, then the resulting belief states can be $\beta_1$, $\beta_3$ or $\beta_5$ with possibility $1_L$, and $\beta_2$ with possibility $0.2$.

The utility function is modified after each iteration :

- Iteration 1 : $u_*(\beta_5) = 1 = \tilde{Q}(\beta_5, S)$ ; $u_*(\beta_3) =$ $0.8 = \tilde{Q}(\beta_3, D)$ ; $u_*(\beta_4) = 0.8 = \tilde{Q}(\beta_4, R)$ ; $u_*(\beta_2) = 0.5 = \tilde{Q}(\beta_2, D)$ ;

- Iteration 2 : $u_*(\beta_2) = 0.8 = \tilde{Q}(\beta_2, D)$ ; $u_*(\beta_1) = 0.5 = \tilde{Q}(\beta_1, R)$ ; $u_*(\beta_7) = 0.5 = \tilde{Q}(\beta_7, R)$ ;

- Iteration 3 : $u_*(\beta_1) = 0.8 = \tilde{Q}(\beta_1, R)$ ; $u_*(\beta_7) = 0.8 = \tilde{Q}(\beta_7, R)$ ; $u_*(\beta_6) = 0.5 = \tilde{Q}(\beta_6, D)$ ; $u_*(\beta_0) = 0.5 = \tilde{Q}(\beta_0, D)$ ;

- Iteration 4 : $u_*(\beta_6) = 0.8 = \tilde{Q}(\beta_6, D)$ ; $u_*(\beta_0) = 0.5 = \tilde{Q}(\beta_0, D)$ ;

Iteration 5 does not change any value, so the algorithm has converged to the following optimal policy (which pessimistic utility is 0.8 whatever the initial state, except if the initial state is $\beta_5$, where its utility is 1) : $(\beta_0, D)$, $(\beta_1, R)$, $(\beta_2, D)$, $(\beta_3, D)$, $(\beta_4, R)$, $(\beta_5, S)$, $(\beta_6, D)$, $(\beta_7, R)$.

## 5 CONCLUSION

In this article we have proposed a qualitative counterpart of POMDPs, based on possibility theory, and in particular on the use of the pessimistic possibilistic decision criterion advocated in [Dubois & Prade1995].

The possibilistic view of multistage decision under uncertainty permits a kind of decomposability of the sequential problem that allows to use dynamic programming-like algorithms for computing optimal policies. In particular, after we described the *possibilistic backwards induction* algorithm proposed in [Fargier, Lang, & Sabbadin1998] for the finite-horizon case, we proposed a modified version, similar to the *value iteration* algorithm, able to cope with problems in which the horizon is not fixed *a priori*.

In order to take partial observability into account in the possibilistic framework, we need to use *conditioning*. Unfortunately, unlike in the stochastic case, there exists no universally accepted notion of conditioning. We chose the one proposed by [Hisdal1978] as it fits well our qualitative (ordinal) framework and it is rather intuitive. Once a possibilistic notion of conditioning is adopted, it is possible to see how belief states (i.e. possibility distributions) are updated after an action is applied and an observation is made. Then, we were able to define the Π-POMDP framework and we showed how Π-POMDPs can be translated into fully observable Π-MDPs over an exponentially larger (but finite) state space.

The algorithm that we proposed for computing optimal policies in the possibilistic partially observable case is based on the possibilistic value iteration algorithm used in the fully observable case. Another approach could be used in which a Π-POMDP is viewed as a game against Nature : trying to maximize the value of the pessimistic criterion amounts to behave like in a game in which Nature chooses the "worst possible" observation, after an action is applied.

Finally, the compatibility of the possibilistic decision criterion with some structured representations of decision problems, particularly in the framework of possibilistic logic should be pointed out. In [Sabbadin1998] it is shown that the possibilistic (pessimistic) one stage decision problem can be stated as an abduction problem with two stratified bases of formulas, one for modeling uncertain knowledge and the other for modeling gradual preferences. Extending this framework to the multi stage, partially observable case, would allow to elaborate a structured language for Π-POMDPs as well as dedicated resolution algorithms.